\documentclass{article}

\usepackage{arxiv}

\usepackage[utf8]{inputenc} 
\usepackage[T1]{fontenc}    
\usepackage{hyperref}       
\usepackage{url}            
\usepackage{booktabs}       
\usepackage{amsfonts}       
\usepackage{nicefrac}       
\usepackage{microtype}      
\usepackage{lipsum}
\usepackage{graphicx}
\usepackage{multirow}
\usepackage{algorithm}
\usepackage{algorithmic}
\usepackage{amsmath}
\usepackage{amssymb}
\DeclareMathOperator*{\argmin}{argmin}
\graphicspath{ {./images/} }

\title{CPOT: Channel Pruning via Optimal Transport}

\author{
  Yucong Shen{\thanks{This work was done when Yucong Shen was a research intern at Tencent AI Lab, China.}}\\
  Department of Computer Science\\
  New Jersey Institute of Technology\\
  Newark, NJ, USA \\
  \texttt{ys496@njit.edu} \\
   \And
  Li Shen {\thanks{Correspondence to: Li Shen <mathshenli@gmail.com> }}\\
  Tencent AI Lab\\
  Shenzhen, China \\
  \texttt{mathshenli@gmail.com} \\
  \And
  Hao-Zhi Huang\\
  Tencent AI Lab\\
  Shenzhen, China\\
  \texttt{matthzhuang@tencent.com} \\
  \And
  Xuan Wang\\
  Tencent AI Lab\\
  Shenzhen, China\\
  \texttt{xwang.cv@gmail.com} \\
  \And
  Wei Liu\\
  Tencent AI Lab\\
  Shenzhen, China\\
  \texttt{wl2223@columbia.edu} \\
}

\begin{document}
\maketitle
\begin{abstract}
Recent advances in deep neural networks (DNNs) lead to tremendously growing network parameters, making the deployments of DNNs on platforms with limited resources extremely difficult. Therefore, various pruning methods have been developed to compress the deep network architectures and accelerate the inference process. Most of the existing channel pruning methods discard the less important filters according to well-designed filter ranking criteria. However, due to the limited interpretability of deep learning models, designing an appropriate ranking criterion to distinguish redundant filters is difficult. To address such a challenging issue, we propose a new technique of Channel Pruning via Optimal Transport, dubbed CPOT. Specifically, we locate the Wasserstein barycenter for channels of each layer in the deep models, which is the mean of a set of probability distributions under the optimal transport metric. Then, we prune the redundant information located by Wasserstein barycenters. At last, we empirically demonstrate that, for classification tasks, CPOT outperforms the state-of-the-art methods on pruning ResNet-20, ResNet-32, ResNet-56, and ResNet-110. Furthermore, we show that the proposed CPOT technique is good at compressing the StarGAN models by pruning in the more difficult case of image-to-image translation tasks.
\end{abstract}


\section{Introduction}
The great progress has recently been made by deep learning techniques at the cost of the requirement of massive computing power. In the context of deploying deep learning model on resource-limited platforms, such a requirement usually cannot be satisfied. In particular, deep CNN models like VGG~\cite{VGG}, ResNet~\cite{resnet} and DenseNet~\cite{densenet} have millions of parameters which demand extensive float-point operations (FLOPs) \cite{FLOPs}. To this end, a variety of model compression approaches have been proposed to reduce the space requirement or/and accelerate the inference process of deep learning models.

In this paper, we focus on pruning methods for deep learning model compression. To convert a certain deep model into a compact one, the network pruning approaches can be categorized into two classes: weight pruning \cite{han2015learning, ding2019global, SNIP, carreira2018learning} and channel pruning \cite{pruneforefficient, SFP, yu2018nisp, FPGM}. In specific, the weight pruning method deletes the specific weights of filters by setting them to zeros, whereas the channel pruning approach deletes the redundant filters (i.e., the channels) entirely. Both approaches, at least in theory, could reduce the model size and speed up the inference. Nonetheless, the resulting sparse convolutional filters by weight pruning are not friendly for implementation. As a consequence, the weight pruning usually achieves model size reduction solely, without model inference acceleration. In contrast, channel pruning first locates the most replaceable filters via a specific criterion and removes them. Then, the channels are reconstructed layer by layer. As are converted to much thinner ones, the channel-pruned deep models become more efficient in both CPU and GPU implementations.

\begin{figure}[t]
\begin{center}
\centerline{\includegraphics[width=3.90in]{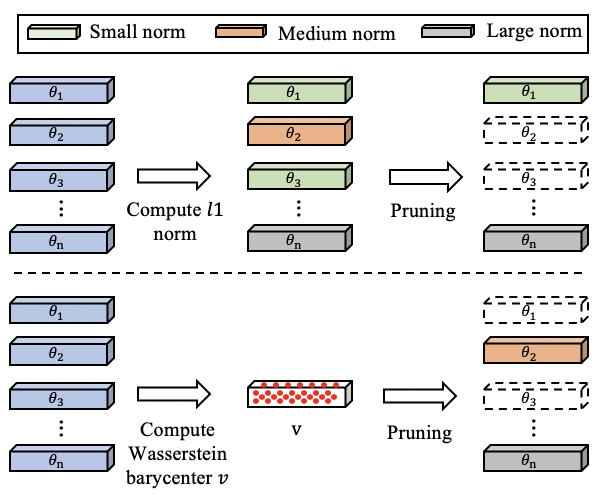}}
\caption{An illustration of the workflow of channel pruning, where the dotted white boxes denote the pruned channels. (\textbf{Top}): The pruning strategy of FPGM \cite{FPGM}, different colors indicate different norms. (\textbf{Bottom}): Our method conducts channel pruning by computing the Wasserstein barycenter $v$ of all the channels, which can be considered as a representation of redundant information.}
\label{pipeline}
\end{center}
\end{figure}

It is obvious that the criteria used to locate the redundant channels are essential for channel pruning. In recent years, researchers have presented several channel pruning approaches with the aim of designing more effective pruning criteria. The methods~\cite{pruneforefficient, ye2018rethinking, SFP} that adopt norm criteria assume that the filters with smaller norms are less important and can thus be discarded. He et al. \cite{FPGM} presented the Filter Pruning via Geometric Median (FPGM) method that prunes filters if they have replaceable distributions. Within each layer of deep models, as shown in Figure~\ref{pipeline}, FPGM particularly calculates the geometric median of filters based on the metric of $\ell_1$ or $\ell_2$ distance and then deletes the filters with relatively small Euclidean distances to the geometric median. Moreover, to capture the group characteristic of channels in the model, it is better for us to perform channel pruning in probability distribution space. Those distances overlook the underlying geometry of the probability distribution space.

\begin{figure}[h]
\centerline{\includegraphics[width=3.90in]{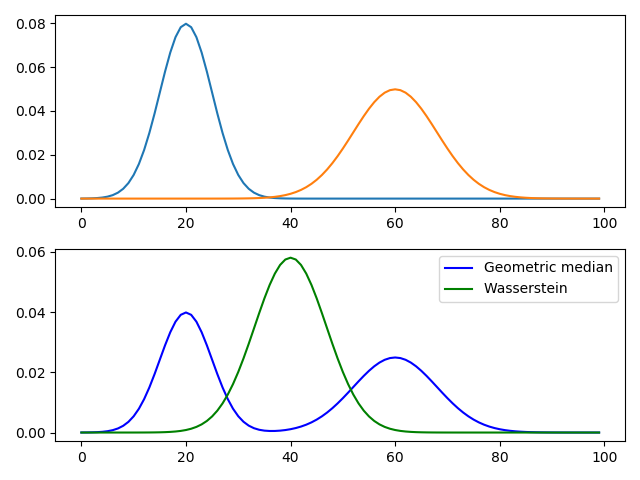}}
\caption{Differences between geometric median and Wasserstein barycenter of two distributions. (\textbf{Top}): Two Gaussian distributions with different means and standard deviations. (\textbf{Bottom}): The geometric median and Wasserstein barycenter of two Gaussian distributions on the top.}
\label{demo_wass}
\end{figure}

To address the aforementioned issue, we adopt optimal transport, which is leveraged to find the optimal transport plan between two probability distributions \cite{yang2019clustering}. Wasserstein distance is defined as the minimum cost of this optimal transport plan, which measures the distance between two probability distributions. Wasserstein barycenter is the mean of a set of probability distributions under the optimal transport metric. This mean minimizes the sum of its Wasserstein distance to each distribution in the set. As shown in Figure~\ref{demo_wass}, Wasserstein barycenter can better capture the geometric characteristics of a set of probability distributions than geometric median.

That motivates us to propose an approach that prunes channels by the tool of optimal transport, which we dub CPOT. By computing Wasserstein barycenter, we can find a better representation of a set of channels. Then the Wasserstein barycenter contains replaceable information that can be represented by other channels. The bottom of Figure~\ref{pipeline} illustrates our idea: we can obtain a representation of all channels by Wassersstein barycenter, making it easy to find the redundant channels. Then channels that have relatively small Wasserstein distances to the Wasserstein barycenter are considered as redundant channels, so they can be regarded as the most replaceable channels. We should be able to remove those channels without largely compromising the model performance. In summary, CPOT discards the channels near the Wasserstein barycenter because they carry redundant information. This process is repeated layer by layer.

Our contributions in this paper are three-fold:
\begin{itemize}
\item We propose a novel channel pruning method of CPOT relying on optimal transport techniques. The proposed method is effective to locate and prune the redundant information by utilizing Wasserstein barycenters of the channels within each layer of deep learning models.

\item We apply the CPOT algorithm to prune ResNets for classification tasks. CPOT outperforms the state-of-the-art methods on several widely-adopted datasets, even for challenging cases that the competing methods cannot solve.

\item We exploit the CPOT algorithm to prune the StarGAN models on more challenging image-to-image translation tasks. Extensive experiments demonstrate that our proposed CPOT can prune starGAN without too much hurting model performance.
\end{itemize}

\section{Related Work}
In this section, we briefly review the literature of pruning approaches for deep learning models, which fall into two major categories: weight pruning and channel pruning.

{\bf Weight Pruning.}\
Many approaches focus on weight pruning. Han et al. \cite{han2015learning} presented a method that eliminates the small weights with a predefined threshold. Lee et al. \cite{SNIP} prune weights based on connectivity sensitivity, which can successfully locate the import connections in neural networks. Ding et al. \cite{ding2019global} proposed to perform weight pruning by a momentum-SGD-based optimization method, which achieved a global compression strategy that automatically determines the sparsity ratio in each layer by end-to-end training. These approaches can dramatically reduce the model size by removing the carefully selected weights from the filters. Nonetheless, it can merely accelerate the inference unless a specific implementation is available.

{\bf Channel Pruning.}\
The approaches of channel pruning locate the redundant channels according to specific criteria and compress deep models by entirely pruning such channels. In other words, channel pruning can inherently reduce the model size and accelerate the inference without any requirement of specific implementations. However, designing proper criteria for channel pruning is a challenging task. In the past few years, some approaches have attempted to tackle this issue, some of which calculate the importance score of each channel. Li et al. \cite{pruneforefficient} pruned filters according to an  $\ell_1$-norm criterion by assuming that filters with small norms have low contributions to neural networks. Yu et al. \cite{yu2018nisp} also created an importance score metric criterion for filter pruning, but it formulates the filter pruning process as binary optimization. It updates the importance score by back-propagating the filter importance scores of the final response layer. The method in~\cite{LCCL} accelerates a convolutional network by leveraging a Low-Cost Collaborative Layer (LCCL), and the method Channel Pruning (CP) \cite{CP} exploits LASSO regression-based channel selection and least-square reconstruction for the same purpose. Very recently, He et al. \cite{FPGM} proposed to prune filters by geometric median which calculates the geometric median of a set of trained filters and then removes the filters that stand near the geometric median in terms of Euclidean distance.

Among the existing channel pruning methods, FPGM \cite{FPGM} is most related to our proposed CPOT algorithm. However, FPGM does not perform well in classifying specific datasets and pruning GAN networks based on our observation. FPGM prunes channels according to the Euclidean distances of channels from their geometric median. Euclidean distance cannot leverage the geometric characteristic of two distributions, and it happens when calculating the geometric median of a group of distributions as well. In contrast, Wasserstein barycenter can better leverage the geometric characteristic of a set of probability distributions. We calculate the Wasserstein barycenter of a set of probability distributions (i.e., channels) based on Wasserstein distance, which can find redundant information in the set. That redundant information can be pruned since it can be represented by the other channels. Therefore, we prune the channels which have relatively small Wasserstein distances to Wasserstein barycenter. In our proposed method, all the channels from each layer are considered as probability vectors under the optimal transport metric.

\section{Channel Pruning via Optimal Transport (CPOT)}

In this section, we describe the proposed CPOT algorithm. We first introduce the methodology of optimal transport. Then the details of our CPOT algorithm are described.

\subsection{Optimal Transport and Barycenter}\label{preliminary}

Optimal transport was first introduced to solve the resources transportation problem \cite{Monge}. Given a set of warehouses and a set of mines, the optimal transport is to transport mines from one warehouse to another in an optimal distance, i.e., Wasserstein distance. Optimal transport has a wide range of applications in clustering \cite{ho2017multilevel}, word embedding \cite{frogner2018learning}, document representation \cite{yurochkin2019hierarchical}, texture analysis \cite{rabin2011wasserstein}, domain adaptation \cite{courty2016optimal}, and GAN models \cite{arjovsky2017wasserstein}.

Since optimal transport can measure the distance between probability distributions very well, we propose our algorithm under the optimal transport metric in the space of probability distributions. Mathematically, given two probability distributions $\mu$ and $\nu$, the minimum cost of optimal transport \cite{xie2018fast}, i.e., $p$-Wasserstein distance, is defined as:
\begin{equation}\label{p-OT}
    W_p(\mu,\nu)\!:=\!\left\{\inf \limits_{\gamma \in \Sigma(\mu, \nu)} \int d^p(x,y)d\gamma(x,y)\right\}^{\frac{1}{p}},\!\!
\end{equation}
where $\Sigma(\mu, \nu)$ is the set of joint distributions whose marginals are $\mu$ and $\nu$, respectively. In the remainder of this paper, we just discuss the case of $2$-Wasserstein distance.

In our algorithm, $\mu$ and $\nu$ are channels, and they are obviously discrete. Therefore, we discuss the above problems in the discrete case. In this case, problem \eqref{p-OT} reduce to the following linear programming \cite{xie2018fast}:
\begin{equation}\label{distrete-OT}
    W(\mathbf{q}, \mathbf{p}) = \min\limits_\Gamma\, \big\langle C, \Gamma \big\rangle,\ s.t.\ \Gamma_i 1 = p, \Gamma^{\top} 1 = q,
\end{equation}
where $\mathbf{q}$ and $\mathbf{p}$ are the given probability distributions.

Moreover, we also adopt the Wasserstein barycenter to acquire the mean of the channels under the optimal transport metric. Mathematically, given a set of probability distributions $\mathcal{P}=\{\mathbf{p_1}, \mathbf{p_2}, \dots, \mathbf{p_n}\}$, their Wasserstein barycenter \cite{xie2018fast} can be defined as:
\begin{equation}\label{barycener}
    \mathbf{q^{*}} (\mathcal{P}, \mathbf{\lambda})= \argmin_{\mathbf{q}\in Q}\sum\limits_{i=1}^n \lambda_i W(\mathbf{q}, \mathbf{p_i}),
\end{equation}
where $Q$ is the space of probability distributions, $\sum_{i=1}^n \lambda_i=1$, and $W(\mathbf{q}, \mathbf{p_i})$ is the Wasserstein distance between the barycenter $\mathbf{q}$ and the probability distribution $p_i$.

Several works have been devoted to solving the optimal transport and Wasserstein barycenter problems \cite{ye2017fast, cuturi2013sinkhorn, benamou2015iterative, ge2019interior, dvurechenskii2018decentralize}. With the sinkhorn algorithm, Benamou et al. \cite{benamou2015iterative} proposed an Iterative Bregman Projections to solve a regularized barycenter in problem \eqref{barycener}.
Recently, Xie et al. \cite{xie2018fast} solved the discrete optimal transport problem \eqref{p-OT} and Wasserstein barycenter problem \eqref{barycener} by developing an Inexact Proximal point method for exact Optimal Transport problem (IPOT).
In this paper, we adopt the IPOT method to seek the Wasserstein barycenters of channels, which are exploited to locate the redundant channels in deep neural networks in the subsequent subsection.

\subsection{CPOT Algorithm}

Now, we are in the position to describe the proposed Channel Pruning via Optimal Transport algorithm. CPOT employs optimal transport to find out and remove the redundant channels in each convolutional layer. In general, our proposed method contains two steps: a) selecting the most replaceable channels within each layer; b) removing them and reconstructing the channels layer by layer.

As shown in Figure~\ref{pipeline}, given a pre-trained deep neural network model, CPOT compresses this model by locating the redundant information in the filters of each layer. Wasserstein barycenter represents the "centroid" of a set of probability distributions \cite{pmlr-v32-cuturi14} under the optimal transport metric. Therefore, the Wasserstein barycenter of a set of probability distributions essentially contains the most replaceable information by the others, and this information can thereby be replaced. That motivates us to locate the redundant information by finding the Wasserstein barycenter of the set of channels in a convolutional layer.

Formally, we use $\theta\in \mathbb{R}^{c\times n\times s_{h}\times s_{w}}$ to represent the filters of the $k$-th convolutional layer. $c$ denotes the input channels, and $n$ is the output channels. Besides, $s_h$, $s_w$ are the height and width of the filters, respectively. When performing channel pruning, we reshape $\theta$ into a matrix $\in \mathbb{R}^{N\times c}$, where $N=n\times s_{h}\times s_{w}$ and each column vector $\in \mathbb{R}^{N\times 1}$ comes from $\theta$. Then our goal is to prune the input channel number from $c$ to $c^{\prime}$, where $c^{\prime}$ should satisfy $0\leq c^{\prime} < c$.

Therefore, in $k$-th layer $\theta$ can be divided into a set of probability vectors (i.e., channels) $\mathcal{W} = \{\omega_1, \omega_2, \dots, \omega_c\}$, where $\omega_i \in \mathbb{R}^{N\times 1}$ is the $i$-th column probability vector in the set. For each set of probability distribution, the Wasserstein barycenter $\mathrm{v}$ can be calculated by solving problem \eqref{barycener}:
\begin{equation}\label{barycenter-CPOT}
    \mathrm{v}(\mathcal{W}, \lambda) = \argmin_{\mathrm{v}\in \mathrm{V}} \sum\limits _{i=1}^c \lambda_i W(\mathrm{v}, \omega_i),
\end{equation}
where $\mathrm{V}$ is the space of the probability distributions.

The Wasserstein barycenter $\mathrm{v}$ is the mean of the set $\mathcal{W}$ under the optimal transport metric. Therefore, after finding the Wasserstein barycenter $\mathrm{v}$ in the $k$-th layer, we successfully find the most replaceable information in the model. As illustrated above, that information can be replaced by the other channels, so we can consider them as redundant information. To this end, we can remove the channels which contain similar information to the Wasserstein barycenter, so that we can compress the model without hurting the model performance too much.

In order to achieve that, we propose to find the probability vectors that are nearest to the Wasserstein barycenter under the optimal transport metric. In Section~\ref{preliminary}, we discuss Wasserstein distance. It measures the distance between two probability distributions. For a set of probability vectors (i.e., channels) $\mathcal{W}$ in the $k$-th layer, to find the probability vectors that are nearest to Wasserstein barycenter $\mathrm{v}$, we have
\begin{equation}\label{CPOT_alg}
    \omega_j = \argmin\limits_ {\omega_{j^{\prime}}} W(\mathrm{v}, \omega_{j^{\prime}}), s.t. j^{\prime}\in\{1,2,\dots, c^{\prime}\}.
\end{equation}
Assume that $\Delta$ is the pre-defined pruning ratio. Then $c^{\prime}=\Delta \cdot c$, and $0\leq \Delta < 1$. $\omega_j$ is the channel to be pruned according to pruning ratio $\Delta$. Since $\omega_j$'s Wasserstein distance to Wasserstein barycenter is smallest in the $k$-th layer. It implies that $\omega_j$ can be represented by the other channels in the convolutional filters of $k$-th layer. Thus, we remove $\omega_j$ by converting all elements of it to zeros to perform channel pruning without compromising the model performance.

Furthermore, in the $k$-th layer, $\theta^{k}$ is updated by back-propagation:
\begin{equation}
    \theta^{(k)}_{t+1} = \theta^{(k)}_{t} - \alpha \frac{\partial J(\theta,b; y, \hat{y})}{\partial \theta^{(k)}}
\end{equation}
at the $t$-th iteration, where $J(\theta,b; y, \hat{y})$ is the objective function, $b$ is the bias, $\hat{y}$ is the target, and $\alpha$ is the learning rate. Specifically, the deviation of the back-propagation begins with chain rule:
\begin{equation}\label{gradient_0}
    \frac{\partial J(\theta,b; y, \hat{y})}{\partial \theta^{(k)}_{ij}} = \frac{\partial J(\theta,b; y, \hat{y})}{\partial a^{(k)}_j} \frac{\partial a^{(k)}_j}{\partial \theta^{(k)}_{ij}},
\end{equation}
where $a^{(k)}_j$ is the output of neuron $j$ in the $k$-th layer before it is passed to the nonlinear activation function. From problem~\eqref{gradient_0}, it is obvious that if $\theta^{(k)}_{ij}$ is equal to zero, there will be a divide-by-zero error. To tackle this issue, we have $\frac{\partial J(\theta,b; y, \hat{y})}{\partial \theta^{(k)}_{ij}}=0$ if $\theta^{(k)}_{ij}$ is in the selected channels. Then we can finish the pruning step of our proposed algorithm. In the end, we summarize the proposed CPOT algorithm in Algorithm~\ref{alg:CPOT}.

\begin{algorithm}[t]
   \caption{\quad CPOT Algorithm}
   \label{alg:CPOT}
\begin{algorithmic}
   \STATE {\bfseries Input:} Training data $X$. Initial model parameters $W=\{W^{(k)}, 0\leq i \leq K\}$.
   \STATE Pre-train the original neural network based on X.
   \STATE Fine-tune the pre-trained model as follows:
   \FOR{$epoch=1$ {\bfseries to} $epoch_{max}$}
   \FOR{$k=1$ {\bfseries to} $K$}
   \STATE Find channels that satisfy Eq. \eqref{CPOT_alg};
   \STATE Convert selected channels to zeros;
   \STATE Convert the gradients of selected channels $\frac{\partial J(\theta,b; y, \hat{y})}{\partial \theta^{(k)}_{ij}}$ to zeros;
   \ENDFOR
   \STATE Update the model parameters $W$ based on X;
   \ENDFOR
   \STATE {\bfseries Output:} The pruned model and its parameters $\theta^{\prime}$.
\end{algorithmic}
\end{algorithm}

{\bf Discussion.} In CPOT, we just perform channel pruning on convolutional layers, and skip fully-connected layers as they are not the main concern of CNN model compression problems. Besides, after we gain the pre-trained model and fine-tune on it, we train the network on the original dataset and prune the whole network iteratively to make sure that the pruned model does not collapse. Since the gradients of selected channels have also been converted to zeros, the neurons from the selected channels will not be updated anymore. After pruning, the compact network can easily resemble the performance of the original model because the pruned channels contain redundant information, which can be represented by the remaining channels. Moreover, as solving the optimal transport and Wasserstein barycenter problem is difficult, which calls for large-scale linear programming, the running times for solving optimal transport and Wasserstein barycenter are $O(n^3)$ and at least $O(n^2 \log n)$ respectively. It is very time-consuming so that a good initial performance of the model is important for our pruning method. This is also the reason why we pre-train the neural network in the very beginning.

\section{Experiments}

We evaluate CPOT's efficacy by compressing different models on different tasks, including multi-branch networks like ResNet \cite{resnet} on CIFAR-10 and CIFAR-100 \cite{cifar} for image classification, and StarGAN \cite{choi2018stargan} on CelebA \cite{celebA} for image-to-image translation.

The CIFAR-10 dataset contains $60,000$ $32\times 32$ color images in $10$ classes. There are $50,000$ training images and $10,000$ testing images. The CIFAR-100 dataset has $60,000$ images in $100$ classes. There are also $50,000$ training images and $10,000$ testing images. CelebA has $202,599$ face images, with $5$ landmark locations and $40$ binary attribute annotations for each image.

Table~\ref{table1} shows the list of datasets, networks, and tasks we use to test our proposed CPOT. Our algorithm is implemented with Pytorch \cite{pytorch}, and tested on $8$ Tesla P40 GPUs.

\begin{table*}[h]
\caption{List of datasets, networks, and tasks that we used to test our CPOT.}
\label{table1}
\begin{center}
\begin{small}
\begin{sc}
\begin{tabular}{lll}
\toprule
Dataset & Task & Networks \\
\midrule
CIFAR-10    & Image classification & ResNet-20, ResNet-32, ResNet-56, ResNet-110 \\
CIFAR-100 & Image classification & ResNet-56, ResNet-110 \\
CelebA    & Image-to-image translation & StarGAN \\
\bottomrule
\end{tabular}
\end{sc}
\end{small}
\end{center}
\vspace{-0.4cm}
\end{table*}

\begin{table*}[t]
\caption{Comparisons of pruning ResNet on CIFAR-10. Baseline acc. shows the model accuracy of the original model, compared methods' baseline acc. come from the original papers. Compressed acc. is the model accuracy after compression. Acc. $\downarrow$ means the model accuracy drops after model compression compared with original baseline accuracy. The smaller, the better. A negative value means that compressed accuracy is higher than baseline accuracy. FLOPs $\downarrow$ is model FLOPs drops after model compression compared with original model FLOPs.}
\label{cifar10}
\vskip 0.15in
\begin{center}
\begin{small}
\begin{sc}
\begin{tabular}{cc|cccc}
\toprule
Model & Method & Baseline acc. (\%) & Compressed acc. (\%) & Acc. $\downarrow$ (\%) & FLOPs $\downarrow(\%)$\\
\midrule
\multirow{3}{*}{ResNet-20} & SFP \cite{SFP} & \textbf{92.20} & 90.83 & 1.37 & 42.2\\
 & FPGM \cite{FPGM} & \textbf{92.20} & 90.44 & 1.76 & \textbf{54.0}\\
 & CPOT & \textbf{92.20} & \textbf{90.88} & \textbf{1.32} &  \textbf{54.0}\\
\midrule
\multirow{4}{*}{ResNet-32} & LCCL \cite{LCCL} & 92.33 & 90.74 & 1.59 & 31.2\\
 & SFP \cite{SFP} & \textbf{92.63} & 92.08 & 0.55 & 41.5\\
 & FPGM \cite{FPGM} & \textbf{92.63} & 91.93 & 0.70 & \textbf{53.2}\\
 & CPOT & \textbf{92.63} & \textbf{92.16} & \textbf{0.47} &  \textbf{53.2}\\
\midrule
\multirow{5}{*}{ResNet-56} & PFEC \cite{pruneforefficient} & 93.04 & 91.31 & 1.75 & 27.6\\
 & CP \cite{CP} & 92.80 & 90.90 & 1.90 & 50.0\\
 & SFP \cite{SFP} & \textbf{93.59} & 92.26 & 1.33 & \textbf{52.6}\\
 & FPGM \cite{FPGM} & \textbf{93.59} & 92.93 & 0.66 & \textbf{52.6}\\
 & CPOT & \textbf{93.59} & \textbf{93.84} & \textbf{-0.25} &  \textbf{52.6}\\
\midrule
\multirow{5}{*}{ResNet-110} & LCCL\cite{LCCL} & 93.63 & 93.44 & 0.19 & 34.2\\
 & PFEC \cite{pruneforefficient} & 93.53 & 92.94 & 0.61 & 38.6\\
 & SFP \cite{SFP} & \textbf{93.68} & 93.38 & 0.30 & 40.8\\
 & FPGM \cite{FPGM} & \textbf{93.68} & \textbf{93.73} & \textbf{-0.05} & \textbf{52.3}\\
 & CPOT & \textbf{93.68} & 93.41 & 0.27 & \textbf{52.3}\\
\bottomrule
\end{tabular}
\end{sc}
\end{small}
\end{center}
\vspace{-0.4cm}
\end{table*}

\begin{table*}[t]
\caption{Comparisons of original FPGM \cite{FPGM} compression results and the CPOT's results of fine-tuning on pruned models obtained by FPGM \cite{FPGM}. Baseline acc. shows the model accuracy without compression. Compressed acc., Acc. $\downarrow$, and FLOPs $\downarrow$ represent the same metric as Table~\ref{cifar10}.}
\label{finetuneGM}
\begin{center}
\begin{small}
\begin{sc}
\begin{tabular}{cc|cccc}
\toprule
Model & Method & Baseline acc. (\%) & Compressed acc. (\%) & Acc. $\downarrow$ (\%) & FLOPs $\downarrow(\%)$\\
\midrule
\multirow{2}{*}{ResNet-20} & FPGM \cite{FPGM} & \textbf{92.20} & 90.44 & 1.76 & \textbf{54.0}\\
 & CPOT & \textbf{92.20} & \textbf{90.99} & \textbf{1.21} & \textbf{54.0}\\
\midrule
\multirow{2}{*}{ResNet-32} & FPGM \cite{FPGM} & \textbf{92.63} & 91.93 & 0.70 & \textbf{53.2}\\
 & CPOT & \textbf{92.63} & \textbf{92.38} & \textbf{0.25} & \textbf{53.2}\\
\midrule
\multirow{2}{*}{ResNet-56} & FPGM \cite{FPGM} & \textbf{93.59} & 92.93 & 0.66 & 52.6\\
 & CPOT & \textbf{93.59} & \textbf{93.00} & \textbf{0.53} & \textbf{63.1}\\
\midrule
\multirow{2}{*}{ResNet-110} & FPGM \cite{FPGM} & \textbf{93.68} & 93.73 & -0.05 & 52.3\\
 & CPOT & \textbf{93.68} & \textbf{94.30} & \textbf{-0.62} & \textbf{74.5}\\
\bottomrule
\end{tabular}
\end{sc}
\end{small}
\end{center}
\vspace{-0.4cm}
\end{table*}

\subsection{Experimental Settings}

\subsubsection{Training Setting}
On CIFAR-10, we iteratively train the model for $200$ epochs with the batch size of $128$ when pre-training. The model is optimized by Adam \cite{adam} with a learning rate of $0.1$. After fine-tuning, we adopt mostly the same training setting as pre-training, but with a learning rate of $0.01$. Besides, we decay $0.0005$ to the original learning rate every $60$ epochs. On CIFAR-100, we use the same setting as on CIFAR-10. On CelebA, we follow the original training and testing settings as in \cite{choi2018stargan}, and we use the same setting as pretraining when fine-tuning the model during channel pruning.

In our observation, fine-tuning from a pre-trained model will endow CPOT with better performance than training from scratch, because the calculation of Wasserstein barycenter is influenced by the initial state. Hence, we only show the experimental results under the fine-tuning after pre-training setting on both the classification task and the image-to-image translation task. Since CPOT is inspired by FPGM \cite{FPGM} and optimal transport, we also fine-tune from the pruned model of FPGM to make a fair comparison.

\subsubsection{Pruning Settings}
In channel pruning, we prune all convolutional layers with the same pruning ratio. After fine-tuning, we prune the channels in each layer after training the model for one epoch. When solving the Wasserstein barycenter and Wasserstein distance problems, $\lambda$ in problem~\eqref{barycenter-CPOT} is set to be $0.01$.

\subsection{Experimental Results on CIFAR-10}
On CIFAR-10, we perform channel pruning on ResNet-20, ResNet-32, ResNet-56, and ResNet-110 by CPOT with the pruning ratio $40\%$.

As shown in Table~\ref{cifar10}, our CPOT achieves better performance than the other state-of-the-art pruning methods. On ResNet-20, the compressed accuracy of CPOT on ResNet-20 is $90.88\%$, which improves $0.05\%$ compared with SFP \cite{SFP}, and prunes $12.0\%$ more FLOPs than SFP \cite{SFP}. On ResNet-32 and ResNet-56, CPOT achieves state-of-the-art performance compared with LCCL \cite{LCCL}, SFP \cite{SFP}, and FPGM \cite{FPGM}. When pruning $53.2\%$ FLOPs on ResNet-32, the accuracy only drops $0.47\%$, which is better than the other methods. On ResNet-56, CPOT's compressed accuracy is $0.25\%$ higher than the baseline, and it prunes $52.6\%$ FLOPs in the meantime. This shows that we successfully locate the redundant information on ResNet-56, and partially solve the over-parameterized problem of deep neural networks on ResNet-56. On ResNet-110, our pruning FLOPs is $52.3\%$, which is better than LCCL \cite{LCCL}, SFP \cite{SFP} and PFEC \cite{pruneforefficient}. However, compared with FPGM \cite{FPGM} and LCCL \cite{LCCL}, our compressed accuracy is a bit lower. Therefore, we show another experiment that fine-tunes and performs pruning on the pruned model obtained by FPGM \cite{FPGM}.

Table~\ref{finetuneGM} shows the results of fine-tuning and pruning on FPGM \cite{FPGM} pruned models by CPOT. On ResNet-20 and ResNet-32, CPOT also achieves better compressed accuracy than FPGM \cite{FPGM}, and keeps the same pruning FLOPs as FPGM \cite{FPGM}. On ResNet-56, although CPOT does not achieve as high compressed accuracy as we showed in Table~\ref{cifar10}, it still beats FPGM \cite{FPGM}, and prunes $63.1\%$ FLOPs, which is much better than FPGM \cite{FPGM} and the other methods in Table~\ref{cifar10}. On ResNet-110, CPOT shows significant improvement. CPOT gets $94.30\%$ compressed accuracy, which is $0.62\%$ higher than the baseline accuracy, and $0.57\%$ higher than the FPGM's \cite{FPGM} compressed accuracy. Besides, CPOT prunes $74.5\%$ FLOPs which is much better than FPGM \cite{FPGM}.

From the experiments on CIFAR-10, compared with FPGM \cite{FPGM}, SFP \cite{SFP}, LCCL \cite{LCCL}, and PFEC \cite{pruneforefficient}, CPOT can accurately locate the redundant information in the deep neural network models. It turns out that optimal transport can find out the relationships between channels in the space of probability distributions.

\subsection{Experimental Results on CIFAR-100}

\begin{table*}[t]
\caption{Comparisons of pruning ResNet on CIFAR-100. Baseline acc., Compressed acc., Acc. $\downarrow$, and FLOPs $\downarrow$ have the same meaning as in Table~\ref{cifar10}.}
\label{cifar100}
\vskip 0.15in
\begin{center}
\begin{small}
\begin{sc}
\begin{tabular}{cc|cccc}
\toprule
Model & Method & Baseline acc. (\%) & Compressed acc. (\%) & Acc. $\downarrow$ (\%) & FLOPs $\downarrow(\%)$\\
\midrule
\multirow{2}{*}{ResNet-56} & FPGM \cite{FPGM} & \textbf{72.42} & 66.93 & 5.49 & \textbf{52.6}\\
 & CPOT & \textbf{72.42} & \textbf{70.88} & \textbf{1.13} & \textbf{52.6}\\
\midrule
\multirow{2}{*}{ResNet-110} & FPGM \cite{FPGM} & \textbf{73.00} & 68.99 & 4.01 & \textbf{52.3}\\
 & CPOT & \textbf{73.00} & \textbf{71.43} & \textbf{1.57} & \textbf{52.3}\\
\bottomrule
\end{tabular}
\end{sc}
\end{small}
\end{center}
\vspace{-0.4cm}
\end{table*}

Since CIFAR-100 has 100 classes, it is more complex than CIFAR-10. To show that CPOT can also deal with more complex image classifications tasks, we show experimental results of CPOT on ResNet-56 and ResNet-110.

In Table~\ref{cifar100}, we show the comparisons between FPGM \cite{FPGM} and CPOT on pruning ResNet-56 and ResNet-110. Comparing to FPGM \cite{FPGM}, CPOT gets better compressed accuracy in the same pruning FLOPs on both of ResNet-56 and ResNet-110. On ResNet-56, the compressed accuracy just drops $1.13\%$ compared with baseline acc., which is even one-fifth of FPGM \cite{FPGM}.

The experimental results on CIFAR-100 show that CPOT achieves better performance than FPGM \cite{FPGM}. With the same pruning FLOPs, CPOT gets much higher compressed accuracy than FPGM \cite{FPGM}. Furthermore, the compressed accuracy does not drop too much compared with the baselines on such a complex dataset. It proves that CPOT can get great performance not only on CIFAR-10, but also on a much more complex dataset. Even when FPGM \cite{FPGM} does not work on CIFAR-100, CPOT can still significantly prune the model without losing too much accuracy. CPOT can gain more meaningful representations by the tool of optimal transport, and prune the redundant information in the deep learning model. It shows that CPOT works for more general datasets, and is suitable for pruning more complex models. In general, we can summarize that CPOT is an excellent choice for pruning the image classification models, as it removes as many channels as possible without affecting model performance too much. CPOT outperforms the other state-of-the-art channel pruning methods. It shows that the  Wasserstein barycenter and optimal transport can efficiently be used to locate redundant information of deep learning models in the space of probability distributions.

\subsection{Experimental Results of Pruning GAN Model}
\begin{table*}[t]
\caption{Experimental results of CPOT for pruning StarGAN \cite{choi2018stargan} on CelebA compared with FPGM \cite{FPGM}. Frechet Inception Distance (FID) \cite{FID} (lower is better) and IS \cite{IS} (higher is better) measure the performances of the models. Baseline FID and Baseline IS are FID and IS of original models. Compressed FID and IS are FID and IS of the models after pruning. Attributes is the numbers of predefined attributes of style transfer.}
\label{GAN}
\begin{center}
\begin{small}
\begin{sc}
\begin{tabular}{cc|cccc}
\toprule
Method & Attributes & Baseline FID & Baseline IS & Compressed FID & Compressed IS\\
\midrule
FPGM \cite{FPGM} & 5 & \textbf{14.20} & \textbf{3.06} & 17.79 & 3.00\\
CPOT & 5 & \textbf{14.20} & \textbf{3.06} & \textbf{17.62} & \textbf{3.04}\\
\midrule
FPGM \cite{FPGM} & 3 & \textbf{12.76} & \textbf{2.99} & 15.22 & 3.01\\
CPOT & 3 & \textbf{12.76} & \textbf{2.99} & \textbf{14.87} & \textbf{3.04}\\
\bottomrule
\end{tabular}
\end{sc}
\end{small}
\end{center}
\vspace{-0.4cm}
\end{table*}
Apart from image classification tasks, we also show that CPOT can work on image-to-image translation tasks. We compress StarGAN's generator \cite{choi2018stargan} by CPOT and FPGM~\cite{FPGM} respectively to compare their performanceS. Our experiments are conducted on CelebA.

In our experiments, the pre-defined pruning ratio is only $30\%$, since GAN models are difficult to compress. The main concern is that the generator would yield implausible images if it discards too many channels.

We pre-train the model for $200,000$ iterations and fine-tune it for additional $200,000$ iterations when performing channel pruning. To avoid making a negative impact on the model performance, we prune the generator every $5,000$ iterations. Two common metrics are adopted to measure the performance of GAN models: Frechet Inception Distance (FID) score \cite{FID} and Inception Score (IS) \cite{IS}. FID \cite{FID} calculates the distance between high-level feature vectors of the generated and real images. It evaluates how similar the generated and real images are. Lower FID scores indicate higher quality of the generated images. Inception score \cite{IS} is also a metric for evaluating the quality of generated images. It is calculated by using a pre-trained Inception-v3 model to predict the class probabilities for each generated image. A higher IS score indicates a higher quality of the generated images.

We show the experimental results in Table~\ref{GAN}. We compress two StarGAN \cite{choi2018stargan} models with 3 and 5 transferred facial attributes. When we are targeting at 5 facial attributes, CPOT can maintain $17.62$ FID  \cite{FID} score after pruning $30\%$ of the channels, which is better than FPGM \cite{FPGM}. IS \cite{IS} also demonstrates the superior performance of CPOT over FPGM \cite{FPGM}. Besides, when we exploit 3 facial attributes, CPOT has $14.87$ compressed FID \cite{FID}, and $3.04$ IS \cite{IS}. CPOT even has higher IS \cite{IS} than the original model. It is clear that CPOT still performs better than FPGM \cite{FPGM}.

Compared with current state-of-the-art methods such as FPGM \cite{FPGM}, CPOT has better FID \cite{FID} and IS \cite{IS} after pruning two generators with different settings. CPOT  keeps decent performances of the image-to-image translation models after channel pruning. As we mentioned above, few pruning methods are proposed to compress GAN models due to the fact that GAN models are more difficult to train than image classification models. As CPOT can also tackle the more difficult GAN models, we conclude that CPOT has a great potential of compressing a wide range of models and datasets. As a result, it demonstrates the effectiveness of the optimal transport theory for channel pruning.

\section{Conclusions}

To make deep learning models be deployed in source-limited devices, various channel pruning methods have been proposed. However, most of them may not be effective enough to compress deep learning models. To this end, we presented a novel algorithm Channel Pruning via Optimal Transport in this paper, dubbed CPOT, to compress deep learning models and generative adversarial networks. Thanks to optimal transport, we are able to locate and prune redundant information in the models under the probability distributions of filters. The experimental results show that CPOT can achieve superior performance over several existing channel pruning methods, such as $\ell_1$ norm pruning and geometry median pruning techniques. Unlike most of the existing work, the proposed CPOT can also be used to prune GAN models for image-to-image translation tasks effectively, such as StarGAN. By comparing against the other existing work, we show that CPOT can be used to compress a range of models on different tasks.

Moreover, there still exist several challenges on CPOT which we leave as future work. For example,  solving optimal transport problems is very time-consuming and taking up a lot of memories on GPU devices. To solve this time-space efficiency issue, we are working on developing a more efficient optimization algorithm to solve it such that large-scale experiments on larger and deeper networks can be tested.

\bibliographystyle{unsrt}
\bibliography{references}

\end{document}